\documentclass[letterpaper, 10 pt, conference]{ieeeconf}  % Comment this line out if you need a4paper

\IEEEoverridecommandlockouts                              % This command is only needed if 
\usepackage{acronym}
\usepackage{subfigure}
\usepackage{graphicx}
\usepackage{float}
\usepackage{subfloat}
\usepackage{caption}
\usepackage{mwe,units}
\usepackage{amsmath}
\usepackage{amsfonts}
\usepackage{xcolor}
\usepackage{array}
\usepackage{hyperref}
\usepackage{url}

\acrodef{EMG}[EMG]{Electromyography}
\acrodef{EEG}[EEG]{electroencephalography}
\acrodef{ECG}[ECG]{electrocardiography}
\acrodef{BMI}[BMI]{Brain-Machine Interface}
\acrodef{SNN}[SNN]{spiking neural network}
\acrodef{AER}[AER]{Address Event Representation}
\acrodef{FPGA}[FPGA]{Field Programmable Gate Array}
\acrodef{PCA}[PCA]{Principle Component Analysis}
\acrodef{SP}[SP]{Separation Property}
\acrodef{ELM}[ELM]{Extreme learning machine}
\acrodef{DPI}[DPI]{Differential Pair Integrator}
\acrodef{SVM}[SVM]{Support Vector Machine}
\acrodef{AFE}[AFE]{Analog Front End}
\acrodef{RMS}[RMS]{Root Mean Square}
\acrodef{CCA}[CCA]{Canonical Correlation Analysis} 
\acrodef{DL}[DL]{Deep Learning}
\acrodef{HMI}[HMI]{Human Machine Interaction} 

\newcommand{\R}{\mathbb{R}}
\usepackage{mathtools,bm}
\DeclareMathOperator*{\argmax}{arg\,max} 

\title{\LARGE \bf
Long-term stable Electromyography classification using Canonical Correlation Analysis
}

\author{Elisa Donati$^{1}$, Simone Benatti$^{2,3}$, Enea Ceolini$^{1}$, and Giacomo Indiveri$^{1}$% <-this % stops a space
\thanks{*This work was not supported by any organization}% <-this % stops a space
\thanks{$^{1}$Elisa Donati, Enea Ceolini, and Giacomo Indiveri are with Institute of Neuroinformatics,
        University of Zurich and ETH Zurich, Winterthurstrasse 190, Zurich, Switzerland
        {\tt\small {elisa, giacomo}@ini.uzh.ch}}%
\thanks{$^{2}$Simone Benatti is with Dipartimento di Scienze e Metodi dell'Ingegneria (DISMI) University of Modena e Reggio Emilia, Modena, Italy,
        {\tt\small {simone.benatti}@unimore.it}}%        
\thanks{$^{3}$Simone Benatti is also with the  Energy Efficient Embedded System (EEES) Lab, University of Bologna, Bologna, BO Italy 40123,
        {\tt\small {simone.benatti}@unibo.it}}%
}

\begin{document}
\maketitle
\thispagestyle{empty}
\pagestyle{empty}

%%%%%%%%%%%%%%%%%%%%%%%%%%%%%%%%%%%%%%%%%%%%%%%%%%%%%%%%%%%%%%%%%%%%
\begin{abstract}
Discrimination of hand gestures based on the decoding of surface electromyography (sEMG) signals is a well-establish approach for controlling prosthetic devices and for Human-Machine Interfaces (HMI).
However, despite the promising results achieved by this approach in well-controlled experimental conditions, its deployment in long-term real-world application scenarios is still hindered by several challenges. One of the most critical challenges is maintaining high EMG data classification performance across multiple days without retraining the decoding system. The drop in performance is mostly due to the high EMG variability caused by electrodes shift, muscle artifacts, fatigue, user adaptation, or skin-electrode interfacing issues.
Here we propose a novel statistical method based on canonical correlation analysis (CCA) that stabilizes EMG classification performance across multiple days for long-term control of prosthetic devices. We show how CCA can dramatically decrease the performance drop of standard classifiers observed across days, by maximizing the correlation among multiple-day acquisition data sets.
Our results show how the performance of a classifier trained on EMG data acquired only of the first day of the experiment maintains 90\% relative accuracy across multiple days, compensating for the EMG data variability that occurs over long-term periods, using the CCA transformation on data obtained from a small number of gestures. This approach eliminates the need for large data sets and multiple or periodic training sessions, which currently hamper the usability of conventional pattern recognition based approaches.
% \indent \textit{Clinical relevance}— Statistical tools to correct the electrode shift and other factors that affect the variability of the EMG signal across days to improve long-term applications.
\end{abstract}

%%%%%%%%%%%%%%%%%%%%%%%%%%%%%%%%%%%%%%%%%%%%%%%%%%%%%%%%%%%%%%%%%%%%%%%%%%%%%%%%
\section{INTRODUCTION}
\label{sec:intro}
Analysis of \ac{EMG} signals is the standard technique to decode the electrical activity of skeletal muscles during their contraction. \ac{EMG} based hand gesture recognition represents a well-established approach for \ac{HMI} in several domains, ranging from robotic control to augmented reality, from personalized medicine to rehabilitation~\cite{Farina2016_surface, Donati2019, Ceolini2020, Zhang2019_multimodal}. 
Conventional approaches are based on mapping \ac{EMG} signal patterns acquired during the muscular contractions, from superficial non-invasive methods, onto a discrete set of gestures. Despite the promising results of many recent basic and applied research efforts, one common and critical challenges that still remains is their robustness over long periods of time.
The reason is that \ac{EMG} signals are highly variable over time, due to many factors such as electrode shifts, muscle fatigue, or skin-electrode interface issues. Typically, within the same experimental session, the \ac{EMG} patterns are repeatable and stable. However, for long-term applications, the donning/doffing of the sensor interface, that may occur during electrodes re-positioning,  dramatically hampers the accuracy of gesture recognition algorithms~\cite{Young2011_effects},~\cite{Milosevic2018_temporal}. Performance drops can reach 30\% figures, making these approaches unsuitable for long-term reliable use~\cite{Milosevic2018_temporal}. 
%The shift is one of the principal reason of the degradation of \ac{EMG} classification accuracy across days~\cite{Young2011_effects}, that can reach a drop of 30\%~\cite{Milosevic2018_temporal}, effecting the long-term usage and the reliability of the prosthesis.

Some solutions have been proposed to overcome this limitation, mostly relying on data set augmentation, including more features~\cite{Kaufmann2010_fluctuating}, all expected displacement locations~\cite{Fan2016_CCA}, multi-session training~\cite{Hu2018}, and Transfer Learning~\cite{Cote2019}.

However, these approaches require long-term training sessions that can be frustrating for the users and consume a considerable amount of power, resulting only in a gain of 10\% of accuracy. Another solution explored to reduce the sensitivity of shift is to use electrodes with larger size, but it has been found that these electrodes perform worse, in terms of gesture classification accuracy, than electrodes with smaller size~\cite{Young2011_effects}.

A promising approach relies on the use of \ac{DL} techniques, that have been shown to be able to learn autonomously the input data representation without having to use handcrafted feature extraction. Hence, these methods can in principle learn input features without being affected by high levels of variability in the data and enable more robust discrimination of the hand gestures. In recent years, \ac{DL} approaches have been successfully deployed for long-term biomedical applications~\cite{Betthauser2019_stable, Tsinganos2019_improved, Zanghieri2019_robust} reaching state-of-the-art accuracy in classification tasks. However, they require very large training data sets. As they are based on complex architectures with very large numbers of hidden layers and parameters, they also require high memory footprint and power consumption which is not always suitable for real-time applications with embedded or portable systems. 

In this work, we tackle the high variability present in a multi-days data set by leveraging a dimensionality reduction technique that maximizes the signal correlation among days. The technique used is the \ac{CCA}, a statistical method that connects two sets of variables by finding a linear combination that maximally correlates them in the mapping space~\cite{Hotelling1936_CCA}. A similar approach, based on \ac{EMG} analysis was already introduced in~\cite{Fan2016_CCA} where two different shift values were investigated to understand the degree of correlation with the normal data; and in~\cite{Khushaba2014} where \ac{CCA} were used to maximize correlation across different users' data. 
Here, we propose to use \ac{CCA} to generalise from one day to an other on the same subject using an \ac{EMG} dataset, collected across 10 days where the electrodes' displacement  --or other sources of temporal variability-- is not known. We propose a framework where we first train a classifier on the data from the first day, and then, we calculate the \ac{CCA} transformation that compensate the variability on a small number of gesture from the next day. This compensation allows using the original classifier in the next days without significant loss of performance, without the need of retraining. 

% In this work we tackle the high variability introduced in a multi-days data set, leveraging a dimensionality reduction technique that try to maximize the correlation between the latent dynamics. We refer to \textit{latent dynamics} as the the time-dependent activation of specific population-wide activity pattern, named \textit{neural modes}, that we hypothesise are stable during generation of the movement. Multiple neural modes define the \textit{neural manifold}, a surface that capture the variance in the recorded activity. The use of these definition is the same proposed in~\cite{Gallego2019_stable} where the authors showed that the latent cortical dynamics are the basic building blocks that underlying a consistent and stable behaviour. The idea behind this work is to investigate if a similar behaviour of latent cortical dynamics is during the generation of movements in the muscles. The muscle activation indeed reflects the activation of the motor neurons in the spinal cord. For this reason, in this paper we are following the same approach proposed for the cortex~\cite{Gallego2019_stable}. We investigate the stability of the latent dynamics within the neural manifolds and their relation across multiple days. The neural modes were computed using low-dimensional reduction such as the PCA.

%%%%%%%%%%%%%%%%%%%%%%%%%%%%%%%%%%%%%%%%%%%%%%%%%%%%%%%%%%%%%%%%%%%%%%%%%%%%%%%%
\section{MATERIALS AND METHODS}
\label{sec:m&m}
%Since one of the final goal of this study is to find a methodology to increase the robustness of \ac{EMG} that is suitable for embedded implementation,
The data set used in this study was acquired in multiple sessions across multiple days set by using a custom \ac{EMG} wearable platform.
%, and used for evaluating the performance of \ac{CCA} and \ac{PCA} based classification algorithms~\cite{salvaro123}.

% \subsection{Acquisition hardware setup}
% The sensor interface is composed of an array of 16 passive gel-based \ac{EMG} electrodes, placed in a ring configuration around the forearm.
% 8 \ac{EMG} channels are acquired at 24-bit resolution in fully-differential configuration by a commercial \ac{AFE} (ADS1298) specially designed for biopotential acquisition. The sampling rate of the ADC ranges from 1 to $32kHz$ and in this application, to minimize the input-referred noise, it is maintained at $4kHz$ The ADC features an internal programmable gain amplifier (1x - 12x) for each channel.  The ADC is controlled via SPI to an ARM Cortex M4 microcontroller, which can stream the data via Bluetooth to an external platform (e.g. a tablet or a benchtop PC). On the board are also present an SD card module to increase the data storage and 2 other sensors (pressure and inertial) to enable data fusion with \ac{EMG} (not used in this experiment)

\subsection{Long-term EMG data acquisition}
\label{subsec:data set}
The sensor interface comprises an array of 16 passive gel-based \ac{EMG} electrodes, placed in a ring configuration around the forearm. 8 \ac{EMG} channels are acquired at 24-bit resolution in fully-differential configuration by a commercial \ac{AFE} (ADS1298) specially designed for biopotential acquisition. The sampling rate of the ADC ranges from 1 to $32kHz$ and in this application, to minimize the input-referred noise, it is maintained at $4kHz$.
To remove the Power Line interference (PLI), we applied a 10-tap Notch filter centered on $50Hz$ frequency. Furthermore, to remove DC wondering and high-frequency noise, we applied a 15-tap Band-Pass filter ($2Hz$-$1kHz$).
The \ac{EMG} data set has been collected over 20 sessions on three able-bodied subjects (all male, average age of 29 $\pm$ 3 years) without neurological disorders~\cite{Zanghieri2019_robust}. 
Data are acquired for 10 days and each day includes two sessions, one in the morning and one in the afternoon. Each session includes 8 hand gestures (palm, fist, index, pinky, hand supination and pronation, ILY sign and thumb up) repeated 8 times with a contraction times of approximately $3s$ with $3s$ rest between two contractions.

\subsection{Multi-day data alignment through \ac{CCA}}
\label{subsec:cca}
To provide a quantitative measure of the correlation between two time series, such as the \ac{EMG} data, we utilized the \ac{CCA}. The \ac{CCA} is deployed for two main aims: i) data reduction, meaning use a small number of linear combinations to explain covariation between two variables; ii) data interpretation that find important features for explaining the covariation between variables. In this work we used \ac{CCA} to find the linear transformation that makes the acquisitions $\mathbf{D}_d$ from day $d$ maximally correlated to $\mathbf{D}_x$ of the reference day $x$, for example in our case the day of the training. 

After the acquisition and segmentation of the \ac{EMG} signals, we extracted features by computing the \ac{RMS} for each channel, a time-domain transformation generally used in \ac{EMG} processing~\cite{Phinyomark2018_feature}. The \ac{RMS} is represented as amplitude relating to a gestural force and muscular contraction and in the implementation proposed is calculated across a window of length $300~ms$ with a sliding window of $100~ms$. Then, for each day, we built a matrix with the same dimension $n \times T$, where $n$ is the number of channels (here $n$ = 8), and $T$ the number of samples from all the concatenated trials of a given day, this includes all of the repetitions for each gesture. 

We indicate with $\mathbf{D}_1$ the matrix assembled from the trials recorded in the day of the training (referring day). We used \ac{CCA} to align the recording in $\mathbf{D}_1$ with those of all the other days using the following approach.
Given two matrices $\mathbf X \in \R^{n \times T}$ and $\mathbf Y \in \R^{n \times T}$ calculated separately for two days, we want to find a linear transformation that makes the corresponding dynamics maximally correlated. In other words we want to find  $\mathbf{A} = \left[\mathbf{a}_1, ..., \mathbf{a}_m\right]$ and $\mathbf{B} = \left[\mathbf{b}_1, ..., \mathbf{b}_m\right]$ with $\mathbf{a}_i$ and $\mathbf{b}_i \in \R^{n}$ so that for $i \in \{1,\dots,m\}$ we have
\begin{align*}
\label{eq:feat1}
    \mathbf {a}_i, \mathbf {b}_i = \argmax_{\mathbf{a} \in \R^{n}, \mathbf{b} \in \R^{n}}Cor(\mathbf{a}^T\mathbf{X}, \mathbf{b}^T\mathbf{Y}) \;\;\;  
\end{align*}

The correlation term can be expanded to reveal the dependence of the maximization problem to the covariances of the data:
\vspace{-0.3cm}
\begin{align*}
    \mathbf {a}_i, \mathbf{b}_i = \argmax_{\substack{\mathbf{a}^T\mathbf{C}_{xx}\mathbf{a}=1 \\\mathbf{b}^T\mathbf{C}_yy\mathbf{b}=1}} \mathbf{a}^T\mathbf{C}_{xy}\mathbf{b} 
\end{align*}

where  $\mathbf{C}_{xy} = \mathbf{XY}^T$, $\mathbf{C}_{xx} = \mathbf{XX}^T$, $\mathbf{C}_{yy} = \mathbf{YY}^T$. 
\\
We can than define a auxiliary variable $\mathbf{\Omega}$ and rewrite the maximization problem 
\begin{align*}
    \mathbf{\Omega} = \mathbf{C}_{xx}^{-1/2} \mathbf{C}_{xy} \mathbf{C}_{yy}^{-1/2} \\
    \mathbf{c} = \mathbf{C}_{xx}^{-1/2}\mathbf{a} \quad \mathbf{d} = \mathbf{C}_{yy}^{-1/2}\mathbf{b}
\end{align*}
\vspace{-0.7cm}
\begin{align*}
     \mathbf{c}_i, \mathbf{d}_i = \argmax_{
     \substack{
     \mathbf{c} \in \R^{n}, \mathbf{d} \in \R^{n} \\ \|\mathbf{c}\|^2=\|\mathbf{d}\|^2=1}
     } \mathbf{c}^T\mathbf{\Omega} \mathbf{d}  
\end{align*}

and 
\begin{align*}
\mathbf{a}_i = \mathbf{C}_{xx}^{-1/2}\mathbf{c}_i \quad \mathbf{b}_i = \mathbf{C}_{yy}^{-1/2}\mathbf{d}_i %\;\;\;  \mbox{for} \; i=1,...,m
\end{align*}

Finally we can show that:
\begin{align*}
SVD(\mathbf{\Omega}) =  \left[\mathbf{c}_1, ..., \mathbf{c}_m \right] \times \mathbf{\Sigma} \times \left[\mathbf{d}_, ..., \mathbf{d}_m\right]
\end{align*}
where $\mathbf{\Sigma}$ contains the canonical correlations.

\subsection{Day to day compensation}
\label{subsec:training}
The principal aim of this framework is to leverage a gesture recognition system that uses \ac{EMG} data measured from the same armband over long periods, across multiple days, but without having to retrain the system during everyday use, after the initial training phase on $\mathbf{D}_1$. To maximize the system's robustness to unknown electrode shifts, a linear transformation is estimated via \ac{CCA} so that this shift can be compensated. This method ensures that the features extracted from the \ac{EMG} signals on days following the referring day can be mapped back to the space of the features used for training the system in the first place. To speed up the process of finding this mapping the transformation is calculated by considering only two repetitions of each gesture. This can be considered as a calibration phase that the end-user can do periodically during normal use of the system. 

The full framework is described as follows: all the data collected on the first day $\mathbf{D}_1$ are used to train ($\sim 6000$ samples for training and $\sim 1500$ sample for testing) a \ac{SVM} to classify the 8 gestures (8 class classification problem) present in the data set. During this phase, it is important to properly regularize the SVM classifier to ensure initial robustness to possible outliers. After training the classifier , the next step is properly adapting it to the data obtained in the following days $\mathbf{D}_x$. First, two repetitions for each gesture are collected. Then, \ac{CCA} is applied to find the optimal transformation between these samples and the corresponding ones obtained during the referring day. As described is Section~\ref{subsec:cca}, this operation returns the mappings $\mathbf{A}$ and $\mathbf{B}$ for the data $\mathbf{D}_1$ and $\mathbf{D}_x$ respectively. To project $\mathbf{D}_x$ back into the space of the features of $\mathbf{D}_1$ the following approach is used   
\begin{equation}
    \hat{\mathbf{D}}_x = (\mathbf{A}^T)^\dagger\mathbf{B}^T\mathbf{D}_x
\end{equation}
where $\dagger$ represents the pseudo-inverse operation (used to avoid numerical instability) and $\hat{\mathbf{D}}_x$ is the projection of the data of the new use day in the space spanned by the features $\mathbf{D}_1$. The same procedure is applied independently for each new day of use. It is worth noting that the proposed framework assumes that the electrodes shift can be compensated by a linear transformation which is a fair assumption that allows for translations and rotation of the electrodes but does not consider changes in the position of the electrodes with respect to each other. 

%%%%%%%%%%%%%%%%%%%%%%%%%%%%%%%%%%%%%%%%%%%%%%%%%%%%%%%%%%%%%%%%%%%%%%%%%%%%%%%%
\section{RESULTS AND DISCUSSIONS}
\label{sec:result}
\begin{figure}
\begin{center}
\includegraphics[width=\columnwidth]{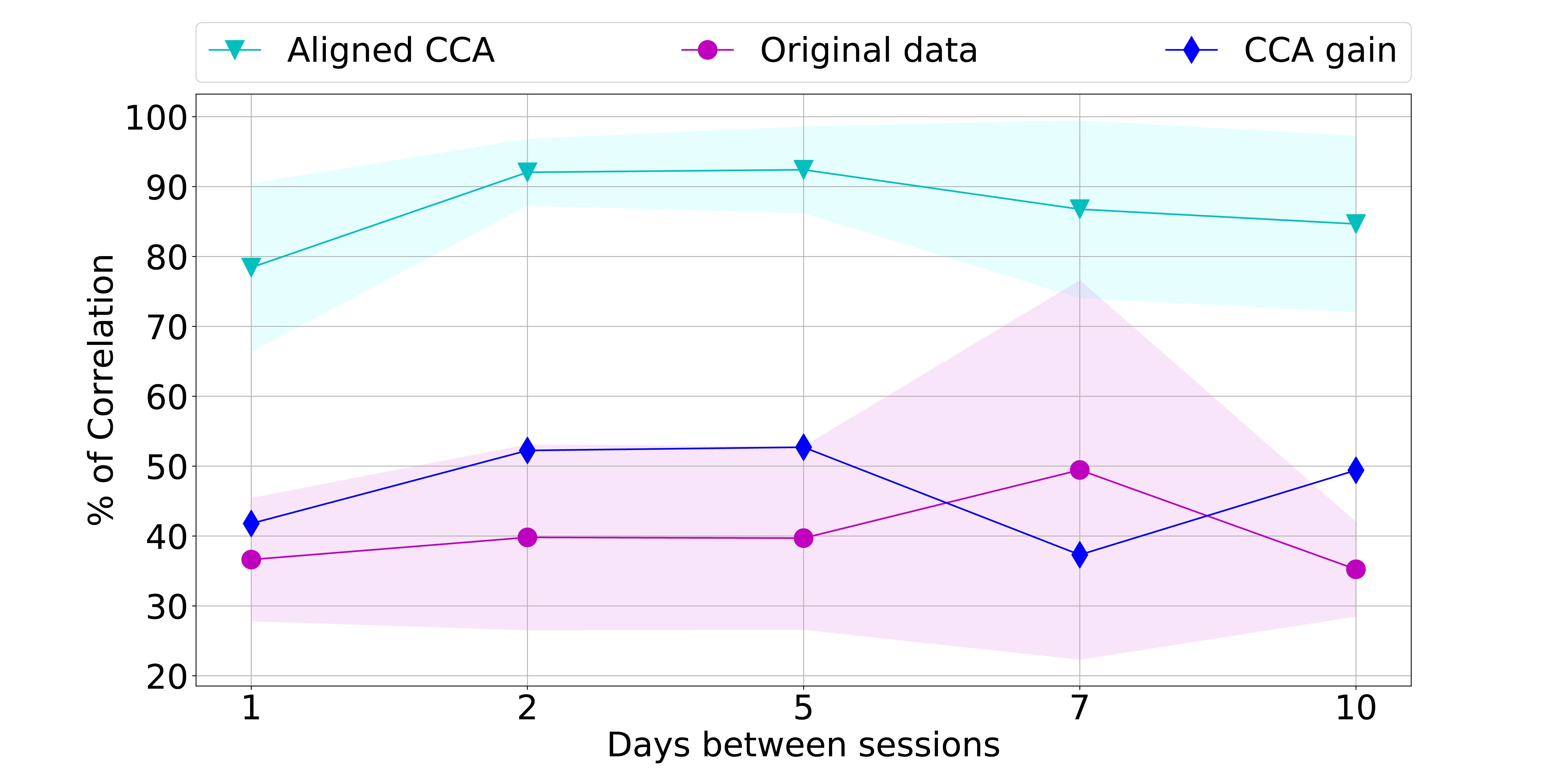}
\caption{Correlation of the \ac{EMG} across the 10 days dataset on three different subjects. Cyan trace represents the aligned correlation across days calculated using \ac{CCA}. The magenta line is the correlation across days without alignment and the blue is the difference between the two correlations.}
\label{fig:Res1}
\end{center}
% \vspace{-5mm}
\end{figure}

Figure~\ref{fig:Res1} shows the correlation between different days. The align system, the one with the \ac{CCA} (cyan) is much more correlated across days than the unaligned one (magenta). We computed normalization on the correlation where the ratio of the across-days aligned with \ac{CCA} has the upper-bound provided by the within-day \ac{CCA}. The blue line indicates the \% of correlation difference between the aligned and unaligned. This corresponds to an average gain in the correlation of about 45\%.  

\begin{figure}
\begin{center}
\includegraphics[width=\columnwidth]{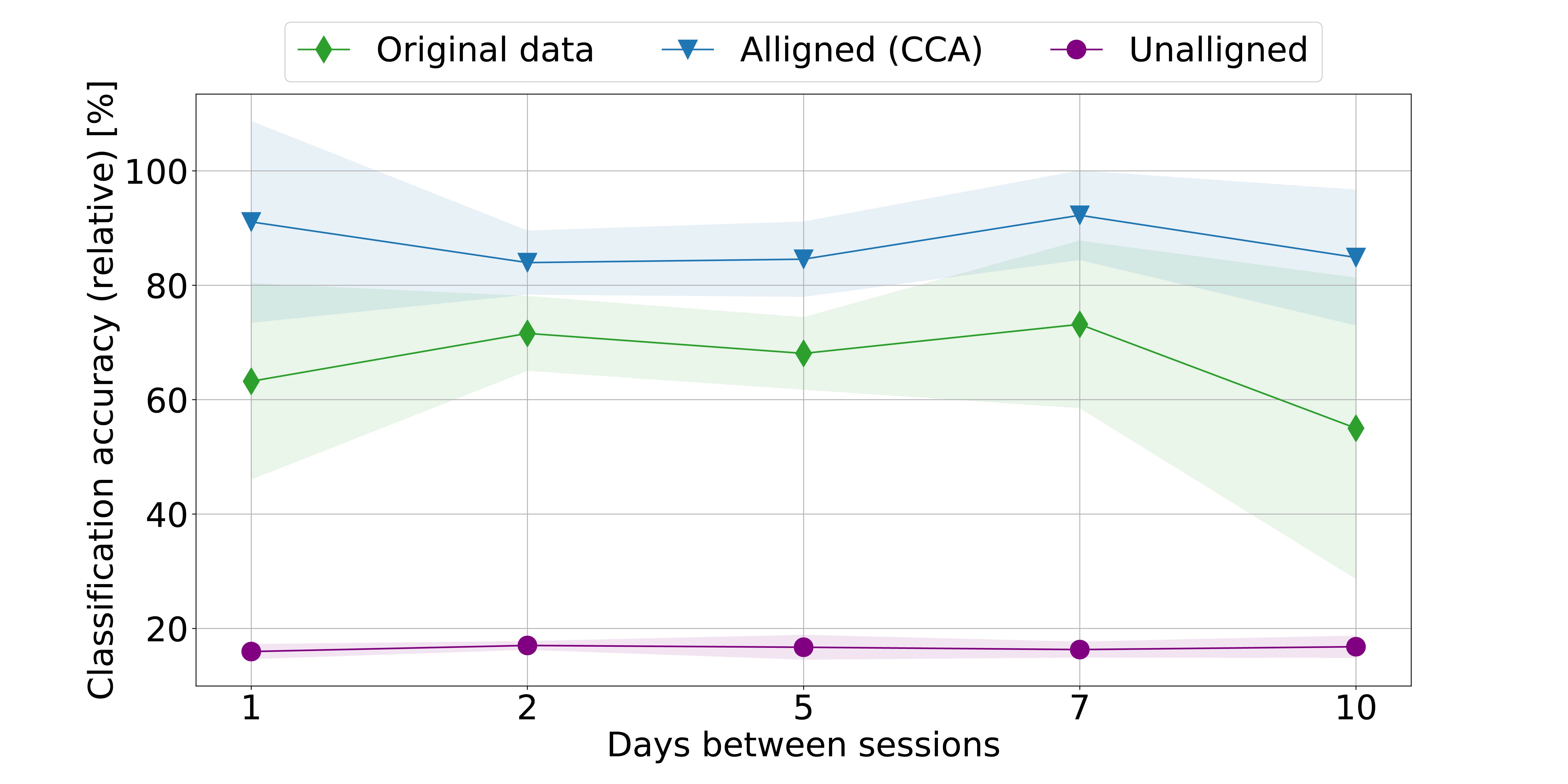}

\caption{Classification accuracy, relative to the accuracy of the classifier on $\mathbf{D}_1$. In purple the classification accuracy for unaligned gestures ($\mathbf{D}_x$). In green the classification on the full dataset and in blue is classification accuracy for gestures aligned with \ac{CCA} ($\hat{\mathbf{D}}_x$)}
\label{fig:Res2}
\end{center}
\vspace{-5mm}
\end{figure}

Figure~\ref{fig:Res2}, shows the classification accuracy of the \ac{SVM} classifier trained on a single day $\mathbf{D}_1$. Without compensation, we can see that the classifier completely fails in classifying the gestures of the other days with an average accuracy across days of around 17\% on a small number of gestures, i.e. two trials for each gesture. The green line represents the classification accuracy on the original full dataset (meaning using samples from all days to train), this results in an average of about 60\%. When compensated with some calibration gestures, the classification reaches almost the same accuracy as if tested on the gestures of $\mathbf {D}_1$. This can be seen by the 95\% relative classification accuracy between the classification of $\mathbf{D}_1$ and the classification of the compensated $\hat{\mathbf{D}}_x$. This shows the effectiveness of calculating a compensating transformation only on several trials. Moreover, this represents a valid alternative to the methods that use re-training of the classifier. The \ac{CCA} compensation can be calculated with a simple one-shot calibration procedure that only needs a limited amount of data, thus avoiding the need for time-consuming and data-hungry re-training that includes new data from the new day.

% embeddings
\begin{figure*}
\centering
\includegraphics[width=\textwidth]{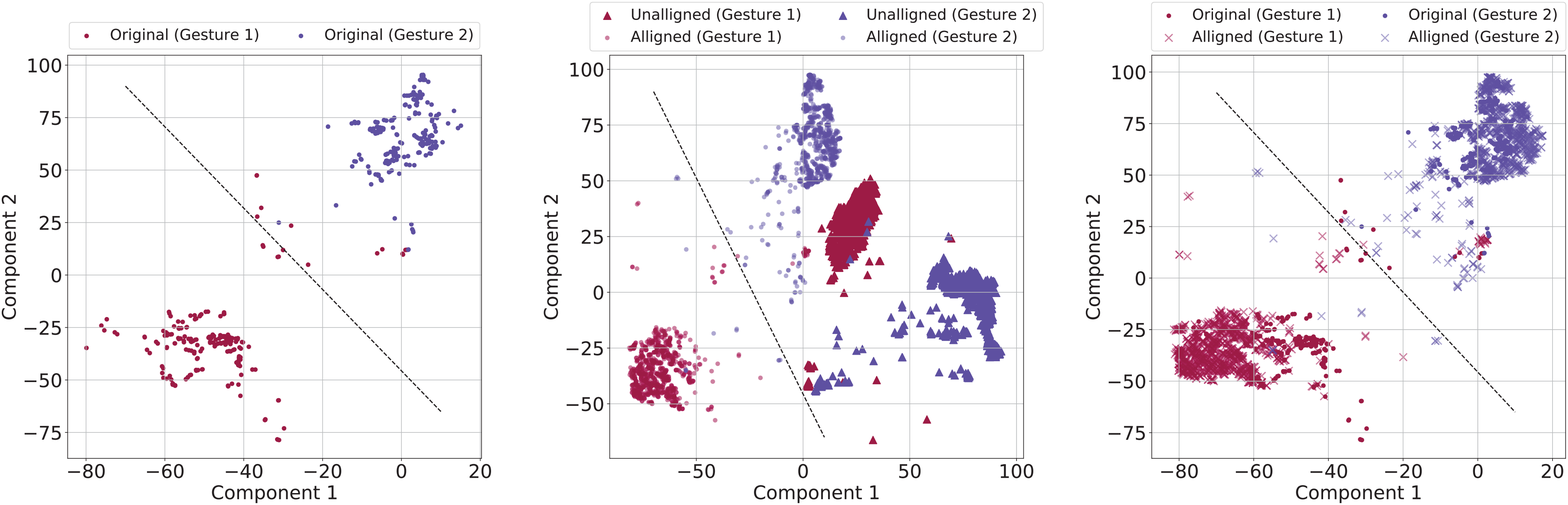}
\caption{Low dimensional projection of two gestures for $\mathbf{D}_1$, $\mathbf{D}_x$ and $\hat{\mathbf{D}}_x$ with the corresponding classification boundary (dashed line). (left) gestures from $\mathbf{D}_1$. (center) gestures from $\mathbf{D}_1$ and $\mathbf{D}_x$. (right) gestures from $\mathbf{D}_1$ and $\hat{\mathbf{D}}_x$}
\label{fig:Res3}

\end{figure*}

Figure~\ref{fig:Res3} shows a low dimensional representation (obtained with t-SNE \cite{Maaten2008tsne}) of two example gestures for $\mathbf{D}_1$, $\mathbf{D}_x$ and $\hat{\mathbf{D}}_x$. In the left panel, we can see that these two example gestures are linearly separable (see separation hyper-plane in dashed line). In the central panel, we see how these two gestures for the $\mathbf{D}_x$ are still linearly separable but they fall within one side of the classifier boundary, showing the reason for such poor classification accuracy in Figure~\ref{fig:Res2}. Notice that the only factor that distinguishes $\mathbf{D}_1$ and $\mathbf{D}_2$ seems to be a rotation, as assumed in the beginning. Indeed, when applying the compensation process we can see in the right panel that the gestures of $\hat{\mathbf{D}}_x$ are now well aligned with the samples of $\mathbf{D}_1$ and are thus correctly classified. 

The overall accuracy across 10 days is around 95\%. If we compare this results with state-of-the-art for regression on the same dataset~\cite{Zanghieri2019_robust} we can see that the accuracy is 95\% vs 93\% obtained with a one-shot calibration instead of an 11-layers (9 convolutional and 2 full) convolutional neural networks trained across the entire dataset. This results in a comparable accuracy but with a significant less computational complexity.

The number of electrodes used in this study was limited to 8. To increase the classification accuracy, e.g., to discriminate between finer gestures, it is possible to use HD-\ac{EMG} systems. The proposed \ac{CCA} analysis can be applied to such high dimensional problems in combination with \ac{PCA} techniques. \ac{PCA} can be performed on the original data to find the time-dependent activation of a specific population-wide activity pattern~\cite{Gallego2019_stable}.
% Once the \ac{PCA} is performed, we would apply the \ac{CCA} on the projected matrix of the PCA to find the maximal correlation between the latent dynamics, that reflect the motor neuron activation.

% , called \textit{latent dynamics}. The use of this definition is the same proposed in where the authors showed that the latent cortical dynamics are the basic building blocks that underlying a consistent and stable behaviour. 
%%%%%%%%%%%%%%%%%%%%%%%%%%%%%%%%%%%%%%%%%%%%%%%%%%%%%%%%%%%%%%%%%%%%%%%%%%%%%%%%
\section{CONCLUSIONS}
\label{sec:conclusion}
In this work, we addressed the temporal variability, in particular, the electrode shift that affects the generalization of the hand gesture recognition based on superficial \ac{EMG}. We show that the \ac{CCA} effectively solves the problem of variability on recordings from the same subject across multi-days recordings and allows a stable and robust gesture classification. In our approach, we trained the classifier on the data from the first day and we compensate the shift in the following days by using the transformation matrix calculate using \ac{CCA} after a calibration phase and not an additional learning. Results show a gain in accuracy in the aligned data of about 30\%, which is even more significant given that we are using a limited amount of data to compensate for the alignment.

\addtolength{\textheight}{-12cm}   % This command serves to balance the column lengths

%%%%%%%%%%%%%%%%%%%%%%%%%%%%%%%%%%%%%%%%%%%%%%%%%%%%%%%%%%%%%%%%%%%%%%%%%%%%%%%%

%%%%%%%%%%%%%%%%%%%%%%%%%%%%%%%%%%%%%%%%%%%%%%%%%%%%%%%%%%%%%%%%%%%%%%%%%%%%%%%%


%%%%%%%%%%%%%%%%%%%%%%%%%%%%%%%%%%%%%%%%%%%%%%%%%%%%%%%%%%%%%%%%%%%%%%%%%%%%%%%%
% \section*{ACKNOWLEDGMENT}
% This work is supported by the EU's H2020 MSC-IF grant NEPSpiNN (Grant No. 753470) and from the EU H2020 research and innovation programme under grant agreement No 826647.

\vspace{-0.2cm}
%%%%%%%%%%%%%%%%%%%%%%%%%%%%%%%%%%%%%%%%%%%%%%%%%%%%%%%%%%%%%%%%%%%%%%%%%%%%%%%%

\end{document}